\documentclass[letterpaper, 10 pt, conference]{ieeeconf}  

\IEEEoverridecommandlockouts   

\makeatletter
\let\NAT@parse\undefined
\makeatother
\usepackage[numbers]{natbib}

\usepackage{amsmath} 
\usepackage{amssymb}  
\usepackage{bm}
\usepackage{xcolor}
\usepackage{graphicx}
\usepackage{mathtools}
\usepackage{hyperref}
\usepackage{cleveref}
\usepackage{siunitx}
\usepackage{tikz}
\usetikzlibrary{shapes}
\usepackage{multirow}
\usepackage{makecell}

\usepackage{subcaption}
\usepackage{tablefootnote}
\usepackage[ruled,vlined]{algorithm2e}

\usepackage{xcolor}
\usepackage[utf8]{inputenc}
\usepackage{pgfplots}
\DeclareUnicodeCharacter{2212}{−}
\usepgfplotslibrary{groupplots,dateplot}
\usetikzlibrary{patterns,shapes.arrows}
\pgfplotsset{compat=newest}
\usepackage{graphicx}
\usetikzlibrary{external}
\usepackage{standalone}
\usepackage{svg}
\Crefformat{figure}{#2Fig.~#1#3}
\Crefmultiformat{figure}{Figs.~#2#1#3}{ and~#2#1#3}{, #2#1#3}{ and~#2#1#3}

\title{\LARGE \bf
MetaLoco: Universal Quadrupedal Locomotion with Meta-Reinforcement Learning and Motion Imitation 
} 

\author{Fatemeh Zargarbashi, Fabrizio Di Giuro, Jin Cheng, Dongho Kang, Bhavya Sukhija, Stelian Coros % 
    \thanks{The authors are with the Computational Robotics Lab, ETH Z{\"u}rich, Switzerland.
    {\tt\footnotesize \{fzargarbashi, fdigiuro, jicheng, kangd, sukhijab, scoros\} @ethz.ch}}%
    \thanks{This research was supported by the European Research Council (ERC) under the European Union's Horizon 2020 research and innovation program grant agreement no.\ 815943 and the Swiss National Science Foundation under NCCR Automation, grant agreement 51NF40 180545.} %
    \thanks{The first two authors contributed equally to this work.} %
    \thanks{The authors thank Yu Zhang for his assistance with the robot experiments.} %
}

\begin{document}

\maketitle
\thispagestyle{empty}
\pagestyle{empty}

\begin{abstract}
This work presents a meta-reinforcement learning approach to develop a universal locomotion control policy capable of zero-shot generalization across diverse quadrupedal platforms. 
The proposed method trains an RL agent equipped with a memory unit to imitate reference motions using a small set of procedurally generated quadruped robots. 
Through comprehensive simulation and real-world hardware experiments, we demonstrate the efficacy of our approach in achieving locomotion across various robots without requiring robot-specific fine-tuning.
Furthermore, we highlight the critical role of the memory unit in enabling generalization, facilitating rapid adaptation to changes in the robot properties, and improving sample efficiency.
\end{abstract}

\section{Introduction}
\begin{figure} 
    \centering
    \includegraphics[width=0.9\linewidth]{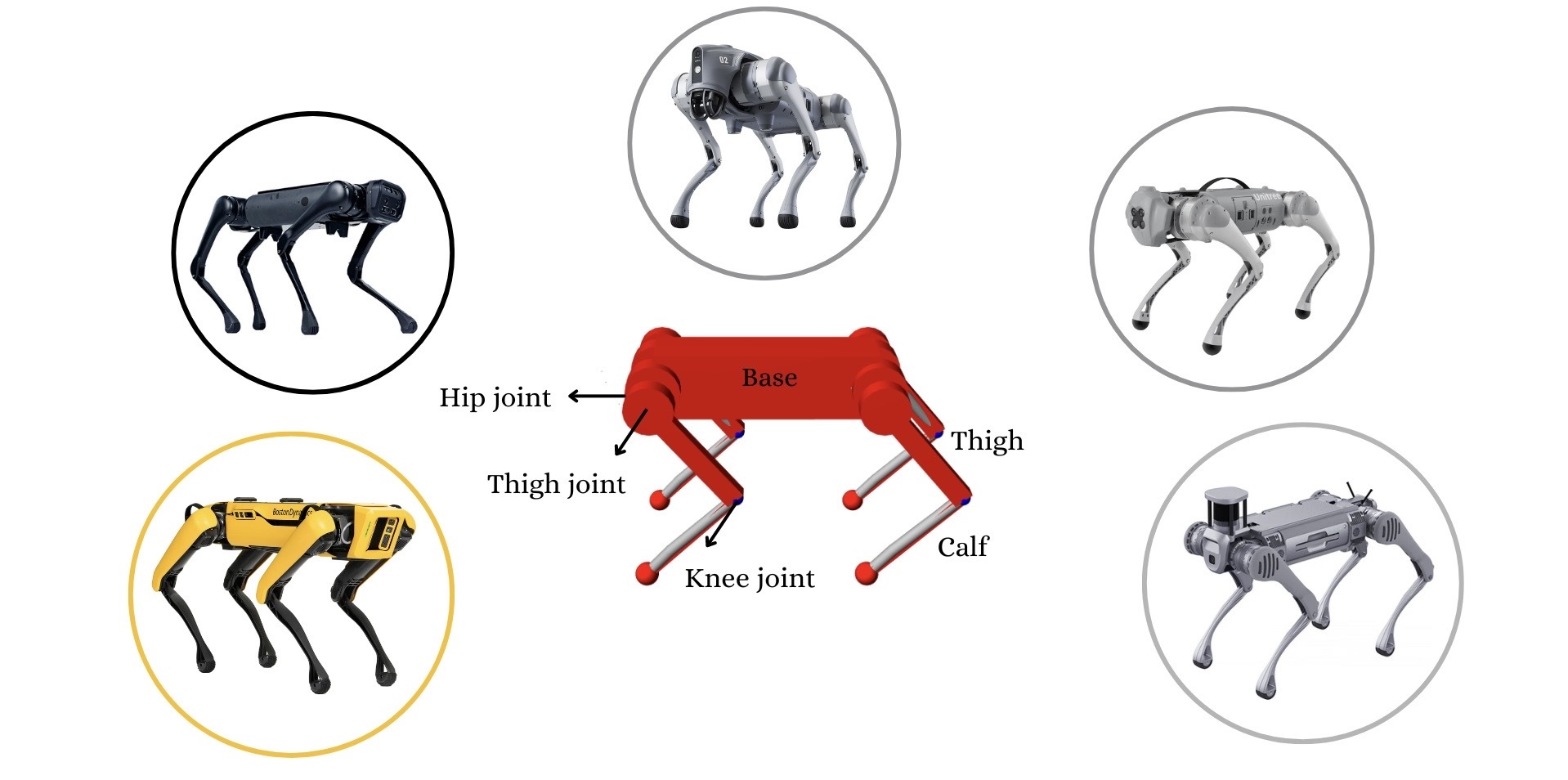}
    \caption{Morphological template of most commercial quadrupedal robots. From left to right: \textit{Boston Dynamics Spot} \cite{spot}, \textit{Unitree Aliengo} \cite{aliengo}, \textit{Unitree Go2} \cite{go2}, \textit{Unitree Go1} \cite{go1}, \textit{Unitree B2} \cite{b2}.} 
    \label{fig: morphology template}
        \vspace{-0.5cm}
\end{figure}

Recent advancements in reinforcement learning (RL) have led to state-of-the-art performance in quadrupedal locomotion \cite{hwangbo2019learning, lee2020learning, miki2022learning, kang2023rl+, choi2023learning}.
Despite their successes, these methods typically develop control policies tailored to specific robot embodiments. 
As shown in \Cref{fig: morphology template}, commercial quadrupedal robots vary widely in design, meaning any changes to a robot's morphology or dynamics often require retraining the policy from scratch.
This retraining process is time-consuming and resource-intensive, often requiring additional reward engineering and hyperparameter tuning, which significantly reduces the practicality and scalability of RL-based locomotion control methods.
To mitigate this challenge, recent research \cite{feng2023genloco, luo2024moral} has shifted focus towards developing a \emph{universal} locomotion control policy capable of \emph{zero-shot} generalization across a wide range of quadruped platforms with varying physical properties. 
Such methods typically involve training on a diverse set of robots to capture a wide range of morphological variations or incorporating an additional system identification module, which increases the complexity of the training and deployment process.
 
In this work, we aim to develop a universal locomotion control policy while enhancing sample efficiency in RL training by limiting the number of robots used during training, thereby improving the practicality of the method.
To achieve this, our method employs \emph{working-memory} meta-RL \cite{duan2016rl, wang2016learning, duan2017meta, rakelly2019efficient}, which provides rapid adaptability and boosts sample efficiency. 
Within this framework, we train a recurrent policy on a relatively small number of quadrupedal robots with distinct morphological and dynamic parameters.
The reward for the RL training is designed to align the robot's states with reference motions from a kinematic motion planner which provides a common motion template while tailoring it to each specific morphology. 
Once trained, the policy operates independently of the planner, enabling seamless transfer across various robot platforms.
Our experiments demonstrate the efficacy of our method in training a universal locomotion policy that generalizes to \emph{unseen} robot embodiments, highlighting the superior sample-efficiency and motion quality achieved by the integration of working memory meta-RL. 
Additionally, we provide insights into how various design choices affect the performance of the proposed method. 
We further validate our method through hardware (HW) tests, where the RL policy successfully achieves zero-shot transfer to three distinct robots: \emph{Unitree Go1} \cite{go1}, \emph{Go2}  \cite{go2} and \emph{Aliengo} \cite{aliengo}, none of which were presented during the training.

\section{Related Work}
\subsection{Reinforcement Learning for Legged Locomotion}
In recent years, RL has significantly advanced legged locomotion by enhancing robustness and reducing manual design effort compared to model-based methods~\cite{kalakrishnan2010fast, bellicoso2018dynamic}, achieving robust locomotion controllers for navigating challenging terrains \cite{hwangbo2019learning, lee2020learning, miki2022learning, choi2023learning}.
However, these methods typically require extensive reward shaping to achieve desirable behaviors.

\emph{Motion imitation} addresses this issue by imitating expert demonstrations or motion data, allowing the RL agent to learn natural and smooth movements more efficiently. 
These demonstrations can be sourced from animal motion clips~\cite{peng2018deepmimic, peng2020learning, Amp_on_hw, yoon2024spatio}, rough hand-held demonstrations~\cite{li2023learning} or optimal control techniques~\cite{kang2023rl+, fuchioka2023opt}.

In this work, we adopt the reference-matching reward structure proposed by \citet{kang2023rl+}, which facilitates learning user-desired behaviors with minimal reward shaping. 
However, unlike the previous work, we train a policy to imitate \emph{kinematic} motion references, rather than model-based optimal control demonstrations.

It is worth noting that RL-based controllers often require a \emph{sim-to-real transfer} to bridge the reality gap---the discrepancies between simulation and real-world performance. 
Researchers employ two primary strategies to address this gap.
One approach involves performing system identification from data collected on the real robot to enhance the simulation fidelity~\cite{hwangbo2019learning, wu2010overview}.
Alternatively, domain randomization is used for mitigating the reality gap by randomizing the physical parameters of the environment during training \cite{peng2020learning, peng2018sim, xie2020learning}.  
To streamline sim-to-real transfer, we also utilize domain randomization, incorporating randomization of parameters such as terrain friction coefficient and actuator latency.

\subsection{Universal Locomotion Control}
With the success of locomotion controllers for quadrupeds, research is increasingly focusing on developing a universal locomotion control applicable to various robot morphologies in a zero- or few-shot manner. 
One research direction integrates robot morphology into the policy architecture.
This involves utilizing architectures that enable generalization across robots with different action and state spaces, such as Graph Neural Networks (GNNs) \cite{wang2018nervenet, huang2020one} or transformers~\cite{gupta2022metamorph,yu2023multi,trabucco2022anymorph}.
However, the validation of these methods has primarily been limited to simulated environments without real-world demonstrations.

From a different perspective, universal control can be approached as a multi-task RL problem, where the policy is conditioned on a vector representation of the robot HW parameters~\cite{chen2018hardware, yu2017preparing, kumar2021rma}. 
This approach is often complemented with a separate module for adaptive system identification \cite{yu2017preparing, kumar2021rma}, and has been successfully applied to quadrupedal locomotion across multiple robot platforms~\cite{luo2024moral}.
More recently, \citet{feng2023genloco} demonstrated that including a history of states and actions into the policy observation can replace explicit HW representation.
This method directly learns to map the history to suitable actions based on inferred kinematic and dynamic properties of the robot.
However, this approach requires training on an extensive number of robot morphologies and has only been demonstrated with a fixed motion, lacking support for velocity commands. 
In contrast, our method aims to develop a universal policy by training on a small number of robot morphologies while being capable of tracking joystick commands.

Universal locomotion control can also benefit from meta-RL algorithms, such as model-agnostic meta learning~\cite{finn2017model}. 
In this approach, a meta-policy is learned across multiple robot designs and subsequently fine-tuned for each specific robot embodiment~\cite{belmonte2022meta}. 
While previous methods focus on few-shot learning, our approach targets zero-shot generalization. 
We redefine universal locomotion control as a working-memory meta-RL problem, training a recurrent policy capable of generalizing across different robots.
Instead of directly inputing the robots' kinematic and physical properties into the policy \cite{belmonte2022meta}, we employ a Gated Recurrent Unit (GRU) \cite{cho2014learning} to infer this information online by updating its hidden state based on the agent's experience in the environment. 

\section{Preliminaries}
\label{sec: meta-reinforcement learning}
Meta-learning, or \emph{learning-to-learn} focuses on developing algorithms that enable models to efficiently learn from experience \cite{vanschoren2018meta}.
Unlike traditional machine learning, where a model is trained for a specific task, meta-learning trains a model to generalize from a set of tasks $\mathcal{T}_{train}=\{{\mathcal{T}_1, \mathcal{T}_2, ..., \mathcal{T}_{N_{train}}}\}$ such that it can quickly adapt to unseen, similar tasks $\mathcal{T}_{test}=\{\mathcal{T}_1, \mathcal{T}_2, ..., \mathcal{T}_{N_{test}}\}$ with minimal or no fine-tuning. 
Effective generalization requires meta-training and meta-testing tasks to share a structure, such as being samples drawn independently and identically from the same distribution, i.e. $\mathcal{T}_{train}, \mathcal{T}_{test} \sim \mathcal{T}$. 

Meta-RL extends meta-learning principles to RL, where each task is a Markov Decision Process (MDP), i.e. $\mathcal{T}_i = \{S_i, A_i, P_i(s_{t+1} | s_t, a_t), R_i(s_t, a_t), \rho_{0,i}\}$. 
Here, $S$ denotes the state set, $A$ the action set,  $P(s_{t+1} | s_t, a_t)$ the transition probability distribution, $R(s_t, a_t)$ the reward function and $\rho_{0}$ the initial state distribution. 
Unlike multi-task RL, which aims to learn a fixed set of tasks, meta-RL leverages experiences from a collection of tasks to cultivate a policy capable of quickly adapting to new tasks~\cite{yu2020meta, beck2023survey}. 
The meta-RL objective is formulated as
\begin{equation}
    \mathcal{J}(\theta) = \mathbb{E}_{\mathcal{T}_i \sim \mathcal{T}}\left[\mathbb{E}_\mathcal{D} \left[ \sum_{\tau \in \mathcal{D}} G(\tau)|f_\theta, \mathcal{T}_i \right] \right] 
\end{equation}
where $G(\tau)$ represents the discounted return in the MDP, $\mathcal{D}$ denotes the meta-trajectory, and $\theta$ are meta-parameters that construct the policy $\pi_\phi$ parameters, $\phi = f_\theta(\mathcal{D})$.

A suitable architectural choice for a meta-learner policy is a Recurrent Neural Network (RNN) architecture, which processes a history of states, actions, and rewards, encoding this information in its hidden state $h_t$.
This method, known as \emph{working-memory meta-RL}, was introduced concurrently by \citet{duan2016rl} and \citet{wang2016learning}. 
Their studies demonstrate that an RNN trained on interrelated environments using a model-free RL algorithm can implement a learning procedure in a new environment through its recurrent dynamics, without the need to adjust network weights~\citep{hochreiter2001learning}. 
However, the adoption of working-memory meta-RL has primarily been limited to simulation due to practical challenges. 
Specifically, including the previous reward into the policy input poses obstacles for HW deployment. 
In this work, we modify working-memory meta-RL as detailed in \Cref{sec: method} to facilitate deployment on HW.
\section{Method}
\label{sec: method}

\subsection{Overview}

\begin{figure*}
    \centering
    \includegraphics[width=0.85\linewidth]{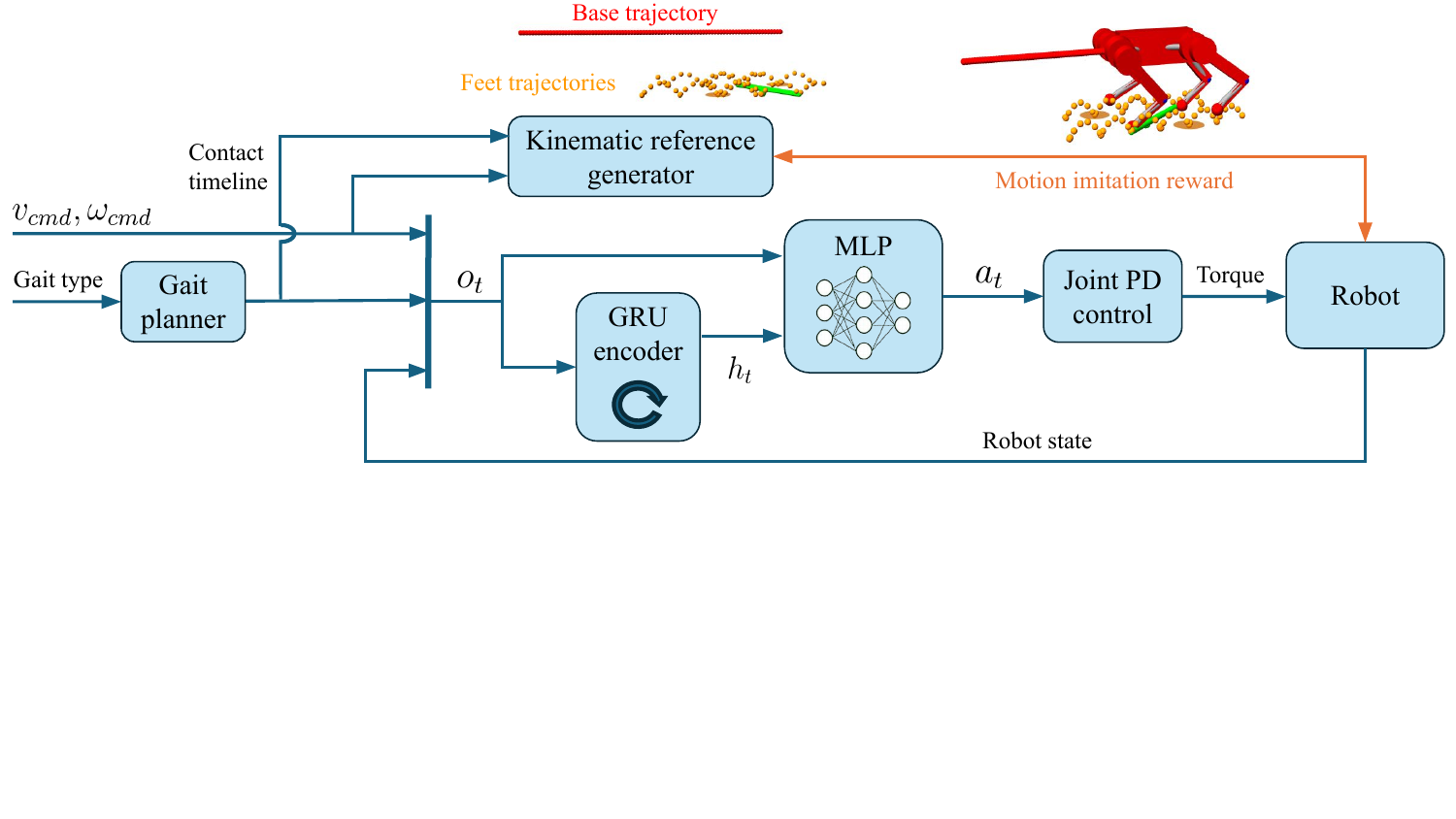}
    \vspace{-3.6cm}
    \caption{Overview of our framework. The objective is to learn a universal policy that given joystick commands, maps the robot’s state ($o_t$) to target joint positions ($a_t = q_t^*$) for various quadruped designs.  The policy is trained to maximise a reward that encourages tracking a reference motion, produced by a kinematic reference generator. }
    \label{fig:method overview}
    \vspace{-0.5cm}
\end{figure*}

Our framework, illustrated in \Cref{fig:method overview}, trains a control policy that determines joint target positions based on the current state and user commands, including gait pattern and velocity command. 
The framework consists of three main components: 1) A gait planner that translates the desired gait pattern into a contact timeline for each leg. 2) A reference motion generator that produces reference trajectories for the base and feet, according to the velocity command, contact timeline and robot properties. 3) A policy trained with meta-RL using motion imitation.

Notably, only the gait planner and the policy are deployed on the robot, while the motion generator is used solely during training. 
This approach makes our control pipeline agnostic to the robot's specific kinematic and dynamic parameters.

We formulate the locomotion task as a motion imitation problem as proposed by \citet{kang2023rl+}. This approach generates reference motions for each episode based on the commanded velocity and gait pattern.
However, instead of employing a model-based optimal planner, we utilize a kinematics planner to produce reference motions.
This planner computes the reference base trajectories and heading angles by numerically integrating the velocity commands.
It further incorporates a straightforward foothold planning rule to generate footholds [\citenum{kang2022animal}, eq. (2,3)].
This enables procedural generation of reference motions for a diverse set of robot embodiment.
Particularly, this approach provides a common motion template shared by every robot embodiment, offering a unified structure in the MDP for each robot by restricting them to share the same motion frequency and pattern.

Inspired by \citet{kang2023rl+}, our reward function comprises an imitation term $r^I$ and a regularization term $r^R$ as
\begin{equation}
r = r^I \cdot r^R = (r^h \cdot r^v \cdot r^{ee} \cdot r^{\dot{\psi}}) \cdot (r^{\Delta a} \cdot r^{slip} \cdot r^{\theta_y, \theta_z}) \ . \label{eq: motion_imitation_reward}
\end{equation}
On the one hand, $r^I$ encourages the alignment of the robot's base height $h \in \mathbb{R}$, velocity $v \in \mathbb{R}^3$, feet position $P_{ee} \in \mathbb{R}^{4 \times 3}$, and yaw rate $\dot{\psi}\in \mathbb{R}$ with the reference trajectories.
On the other hand, $r^R$ regulates action rate $\Delta a \in \mathbb{R}^{12}$, while minimizing contact feet velocity $v_{ee} \in \mathbb{R}^{4 \times 3}$, base pitch $\theta_y \in \mathbb{R}$ and roll $\theta_z \in \mathbb{R}$ angles. 
Each reward term maps the error between the reference $\mathbf{x}$ and the actual value $\hat{\mathbf{x}}$ to a scalar $r^x \in [0, 1]$ using the radial basis function kernel, with sensitivity $\sigma_x$:

\begin{equation}
r^x = \exp\left(-\left\lVert \hat{\mathbf{x}} - \mathbf{x}\right\rVert^2 / \sigma_x^2\right). \label{eq: RBF kernel}
\end{equation}

\subsection{Meta-RL Setup}
In our problem formulation, each task $\mathcal{T}_i$ represents a motion imitation task for a quadruped with embodiment $i$. 
Here, all tasks share a common state and action space, but differ in their dynamics $P_i$ and initial state distribution $\rho_{0,i}$. 
The reward functions share a similar structure but are task-dependent, as the imitation reference varies according to the robot properties.

\begin{figure*}[tb]
    \centering
    \vspace{0.1cm}
    \includegraphics[width=0.95\linewidth]{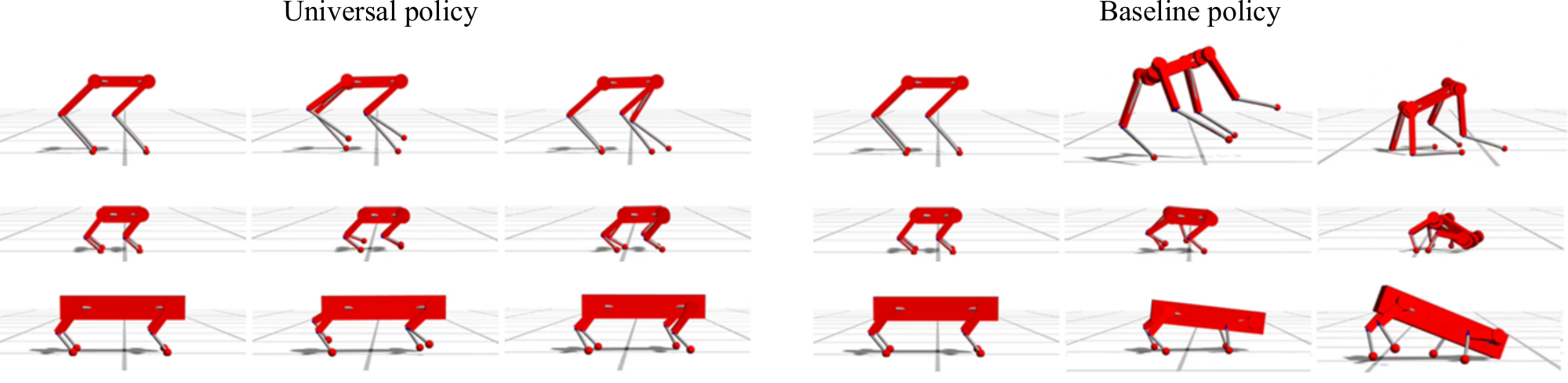}
    \caption{Three simulated quadrupeds successfully trotting with our universal policy (\textbf{left}) and failing with a policy specifically trained for Go1 (\textbf{right}).}
    \label{fig: snapshots of controllable morphologies}
    \vspace{-0.5cm}
\end{figure*}

The policy architecture comprises a GRU with a hidden state of size 16, which processes a sequence of observations. 
The hidden state $h_t$ is concatenated with the current observation $o_t$ and passed as input to a multi-layer perceptron (MLP), as illustrated in \Cref{fig:method overview}, with a linear output layer applied afterwards.
In the motion imitation setup, computing the reward during HW deployment requires a reference, which in turn necessitates knowledge of the robot's kinematic and dynamic parameters.
This conflicts with our goal of achieving quadruped-agnostic control. 
Therefore we modify the meta-learning method by dropping the reward from inputs to our policies.

The policy is trained using a modified Proximal Policy Optimization (PPO) \cite{schulman2017proximal}, tailored for working-memory meta-RL. 
In this variant, fixed-length sequences of transitions are sampled from the rollout buffer at each training step, rather than single transitions. 
Furthermore, following the meta-episodic framework proposed by Duan et al. \cite{duan2016rl}, during rollout collection, the GRU hidden state is preserved across episodes, with a periodic reset every $K$ episodes. 
This approach allows the agent to accumulate and leverage information about the task structure and dynamics across episodes. An analysis of the impact of different choices of $K$ is provided in \Cref{result:meta-episode}.

\subsection{Implementation Details}

The observation of the policy $o_t$ includes proprioceptive information, namely joint position $q$ and speed $\dot{q}$, base height $h$, base linear velocity $v$ and angular velocity $\omega$, gravity vector $g$, as well as joystick commands including forward velocity $v_{cmd}$, turning velocity $\omega_{cmd}$, gait commands expressed as contact phase variables $\cos(\phi)$, $\sin(\phi)$, and finally, the previous action $a_{t-1}$. 
We use the contact phase parameterization proposed by \citet{shao2021learning}, which represents a swing phase as $\phi \in [-\pi, 0)$ and a stance phase as $\phi \in [0, \pi)$. 
The action $a_t$ is the joint position targets, which are tracked by a PD controller.
The policy is queried at a rate of 50 Hz, whereas the low-level PD controllers run at 200 Hz. 
Details of the reward and training hyperparameters are provided in \Cref{tab: Reward Hyperparameters} and \Cref{tab: PPO Hyperparametes}, respectively.

\begin{table}[b]
\vspace{-0.2cm}
\caption{Reward hyperparameters}
\vspace{-0.2cm}
\begin{center}
\begin{tabular}{|c|c||c|c|}
\hline
\textbf{Reward term}
& \boldsymbol{$\sigma_x$} & \textbf{Reward term} & \boldsymbol{$\sigma_x$} \\
\hline
     base height $r^h$ & 0.05 &
     base velocity* $r^v$ & [0.3, 0.1, 0.3] \\
     base yaw rate $r^{\dot{\psi}}$ & 0.5 &
     feet position* $r^{ee}$ & [0.3, 0.05, 0.3] \\
     action rate $r^{\Delta a}$ & 1.5 &
     feet slip $r^{SLIP}$ & 0.1 \\
     pitch and roll $r^{\phi, \theta}$ & 0.5  &  &\\     
\hline
\multicolumn{4}{l}{$^{*}$ Non-scalar values are applied in forward-vertical-sideways order.}
\end{tabular}
\label{tab: Reward Hyperparameters}
\end{center}
\vspace{-0.4cm}
\end{table}

\begin{table}[bhtp]
\caption{Training hyperparameters}
\vspace{-0.2cm}
\begin{center}
\begin{tabular}{|c|c||c|c|}
\hline
\textbf{Hyperparameter} & \textbf{Value} & \textbf{Hyperparameter} & \textbf{Value} \\
\hline
     Batch size & 512 &
     Number of epochs & 10 \\
     Value coefficient & 0.5 &
     Entropy coefficient & 0.01 \\
     Discount factor & 0.95 &
     Learning rate & $5 \times 10^{-5}$ \\
     Episode length & 128 &
     Initial standard deviation & $\exp(-1)$ \\
     Sequence length & 16 &
     Meta-episodic length & 5 \\      
\hline
\end{tabular}
\vspace{-0.3cm}
\label{tab: PPO Hyperparametes}
\end{center}
\end{table}

\begin{figure}
    \centering
    \includegraphics[width=0.95\linewidth]{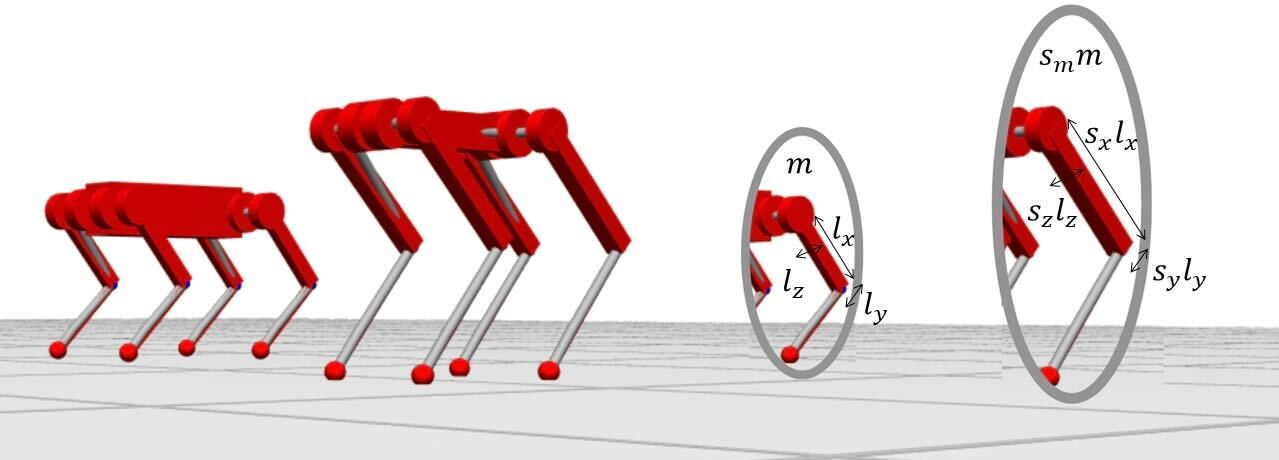}
    \caption{Visualization of the morphology structure of  \textit{Unitree Go1} (\textbf{left}), and one generated with our method (\textbf{right}). In the ovals, we highlight the parameters of the thigh link of the hind-left leg.} 
    \label{fig: morphology generation}
    \vspace{-0.6cm}
\end{figure}

At the start of each training episode, we randomly sample a velocity command and the initial robot state from one of the reference motions, a technique known as reference state initialization (RSI)~\cite{peng2018deepmimic}. 
We terminate an episode early if the robot collapses, i.e. if any non-foot part of the robot comes into contact with the ground.

\subsection{Morphology Generation}
\label{sec: morphology generation}

\begin{figure}
    \centering
    \includegraphics[width=0.95\linewidth]{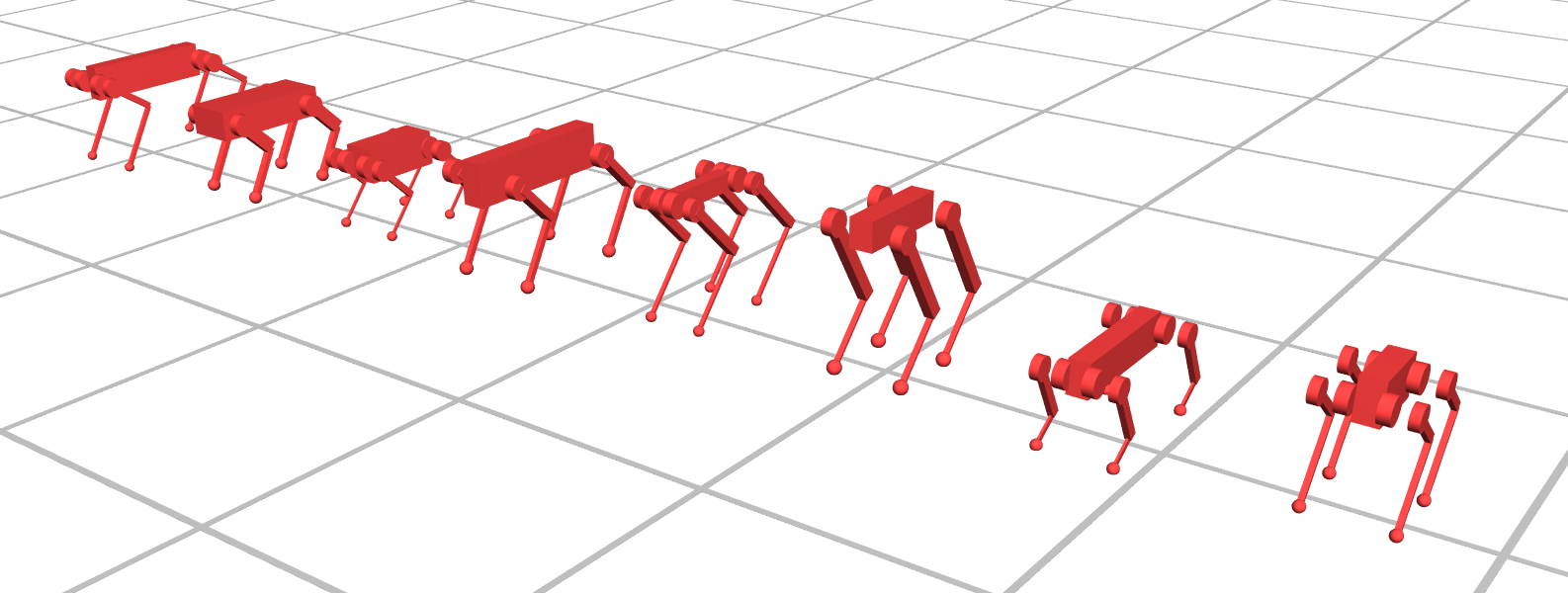}
    \caption{Example of quadrupeds procedurally generated by randomizing the kinematic and dynamic parameters of \textit{Unitree Go1} and \textit{Unitree Aliengo}.} 
    \label{fig:procedural_morphs}
    \vspace{-0.5cm}
\end{figure}

Commercially available quadrupedal robots typically follow a design template (see \Cref{fig: morphology template}) comprising an unactuated base and four 3-DOF legs. Each leg is composed of a hip, a thigh, and a calf link, terminating in a point-foot.
Given the morphology of a commercial robot, we generate a new morphology by updating the dimensions and mass properties of each link as illustrated in \Cref{fig: morphology generation}. For each link type (base, calf, thigh), we sample four scaling factors $[s_x, s_y, s_z, s_m]$ independently from a uniform distribution U$[0.5,1.5]$. These scaling factors are applied to the nominal dimensions $[l_x, l_y, l_z]$ and mass $m$ of all instances of the link type.
The matrix of inertia $I$ of each link is recalculated by approximating the link with a cube:
\begin{equation}
I = 
\begin{bmatrix}
\frac{1}{12}m(l_y^2+l_z^2) & m(l_x \cdot l_y) & m(l_x \cdot l_z)\\
m(l_y \cdot l_x) & \frac{1}{12}m(l_x^2+l_z^2) & m(l_y \cdot l_z)\\
m(l_z \cdot l_x) & m(l_z \cdot l_y) & \frac{1}{12}m(l_x^2+l_y^2) 
\end{bmatrix}.
\label{eq:moi_formula}
\end{equation} 
Additionally, we adjust the robot thigh joint angle such that every foot lies below the corresponding hip joint, as required for stability and locomotion efficiency.
Some of the generated morphologies are shown in \Cref{fig:procedural_morphs}

\section{Simulation results}
\label{sec:sim}

To verify the efficacy our method, we conducted a series of comprehensive simulation experiments.
All policies employed in the following experiments were trained using our framework on a set of procedurally generated robots, as detailed in \Cref{sec: morphology generation}. 
To ensure a broad spectrum of training morphologies, we generated the robot set by randomizing parameters of \textit{Unitree Go1} and \textit{Aliengo}.
Each policy was trained to imitate a trotting gait reference with a gait cycle of \SI{0.5}{\second}, under varying velocity commands ranging from \SI{-0.5}{\meter/\second} to \SI{1.0}{\meter/\second}.
We utilized Open Dynamics Engine (ODE) as the physics simulator~\cite{smith2005open}. 
All policies were trained for 163.84 million samples \emph{without} dynamics randomization, apart from exposure to the set of different embodiments.
Notably, our method is significantly more sample efficient compared to the approach in ~\citet{feng2023genloco}, which requires 800 million samples for training.

\subsection{Generalization over diverse embodiments}
We demonstrate the efficacy of our  universal control policy by testing it on a diverse population of quadrupeds with varied morphological and dynamic properties, which were \emph{unseen} during training.   
For comparison, we also trained a baseline policy with a standard RL training setup without randomized embodiments
and without a recurrent cell. 
Qualitative evaluations on a test set of 40 robots demonstrates that our policy effectively generalized to unseen robots while keeping the trotting gait pattern, maintaining the commanded speed, and achieving target values for base height and foot swing height as illustrated in \Cref{fig: snapshots of controllable morphologies}.  
In contrast, the baseline policy failed to control the unseen morphologies.

\subsection{Comparison of Architectures}

\begin{figure}
    \centering
    \includegraphics[width=0.9\linewidth]{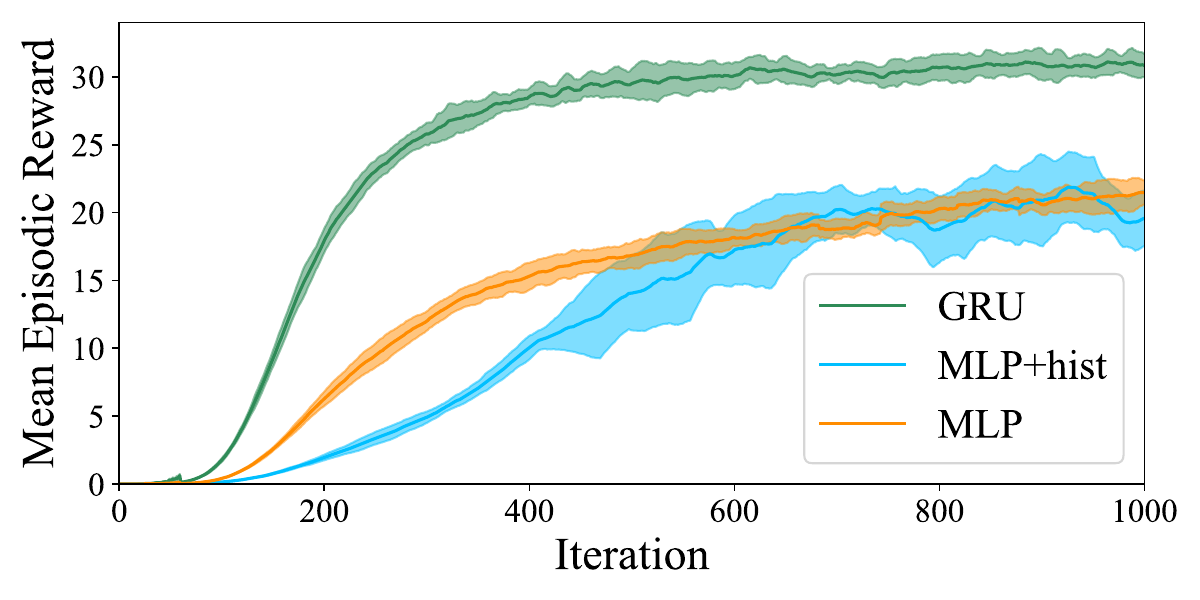}
    \caption{Mean and standard deviation of reward curves during training for different architectures, over 5 different seeds.
    } 
   \label{fig: tikz learning curves}
   \vspace{-0.4cm}
\end{figure}

To demonstrate the critical role of the GRU encoder, we compared our policy architecture against two other baselines: \emph{MLP} and \emph{MLP+history}. The latter, as used by \citet{feng2023genloco}, takes as input a history of the last 16 observations.
For each architecture, we trained 5 distinct policies using different random seeds.
Each policy was trained on the same set of 32 robots.
The training process took approximately 17, 20 and 28 hours on an NVIDIA GeForce RTX 2080 Ti for \emph{MLP}, \emph{MLP+history} and \emph{GRU} policies, respectively.

As depicted by the learning curves in \Cref{fig: tikz learning curves}, GRU policies attain a significantly higher final reward and achieve better sample efficiency compared to other network architectures. 
Interestingly, providing a history of observations to the MLP does not improve the learning curve; instead, it increases variance. 
This could be attributed to constraints in network capacity and mitigated by utilizing a larger network.

We conducted two robustness experiments across a test set of 40 \emph{unseen} quadrupeds and a total of 2000 episodes:

1) \emph{Walk test} involves walking for  $4 s$ with randomized speed commands $v_{cmd}$. 

2) \emph{Mass curriculum test} involves walking for $8 s$ with speed commands $v_{cmd} \in \{-0.3, 0.3\} \SI{}{\meter/\second}$.
The trunk mass is abruptly scaled by a factor of $s_m$ every \SI{2}{\second}.

The Mean Episodic Reward (MER) served as the performance metric, computed as the average cumulative reward across all episodes:
\begin{equation}
    MER = \frac{1}{N_{\text{total}}} \sum_{i=1}^{N_{\text{total}}} \left( \sum_{t=0}^{H} r_{i,t} \right).
\end{equation}

\begin{table}
\vspace{0.1cm}
\caption{Walk experiment: Mean Episodic Reward (MER)}
\vspace{-0.2cm}
\begin{center}
\begin{tabular}{|c|c|c|c|}
\hline
$v_{cmd} \ [m/s]$ & GRU & MLP + hist & MLP \\
\hline
    $\{-0.1, 0.1\}$ & \textbf{103.761} & 95.480 & 99.160 \\
    $\{-0.2, 0.2\}$ & \textbf{99.520} & 89.564 & 95.886 \\
    $\{-0.3, 0.3\}$ & \textbf{92.162} & 81.708 & 90.764 \\
    $\{-0.4, 0.4\}$ & \textbf{83.048} & 72.413 & 82.610 \\
\hline
\multicolumn{4}{l}{}
\end{tabular}
\label{tab: MER for 32 robots (walk)}
\end{center}
\vspace{-0.6cm}
\end{table}
\begin{table}
\caption{Mass curriculum experiment: Mean Episodic Reward (MER)}
\vspace{-0.2cm}
\begin{center}
\begin{tabular}{|c|c|c|c|}
\hline
$s_{m}$ & GRU & MLP + hist & MLP \\
\hline
$0.8$ & \textbf{183.761} & 164.648 & 178.338\\
$0.9$ & \textbf{187.313} & 167.127 & 182.313\\
$1.1$ & \textbf{185.936} & 166.120 & 179.579\\
$1.2$ & \textbf{172.843} & 165.581 & 170.092\\
\hline
\multicolumn{4}{l}{}
\end{tabular}
\label{tab: MER for 32 robots (mass curriculum)}
\end{center}
\vspace{-0.9cm}
\end{table}
The results in \Cref{tab: MER for 32 robots (walk)} and \Cref{tab: MER for 32 robots (mass curriculum)} indicate that our GRU policy outperforms both MLP variants, underscoring the importance of the GRU encoder for generalizing to unseen robots and adapting to dynamic parameter variations never seen during training.

\begin{figure}[t]
    \centering 
        
    \includegraphics[width=0.9\linewidth]{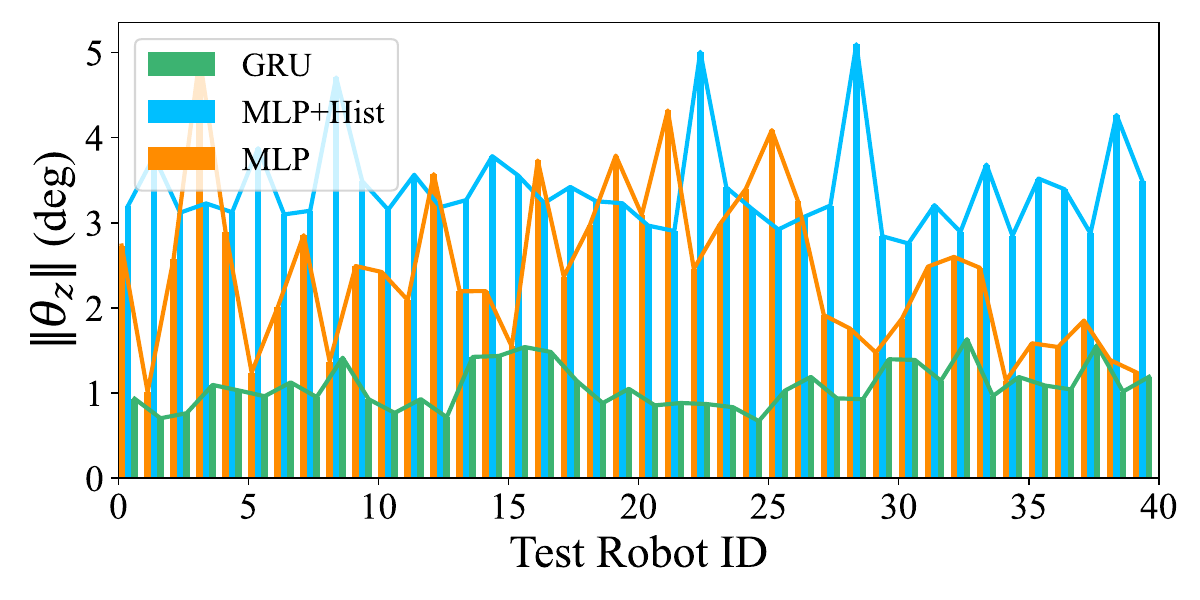}        \vspace{-0.2cm}
    \caption{Quality of motion experiment: root mean squared error of roll angle $\|\theta_z\|$ for different architectural choices, across the test robots.}
    \label{fig:tikz_roll_angle}
\end{figure}

Additionally, we visually compared the quality of the motion induced by different architectural choices, as shown in the supplementary video. 
Notably, only the policy with the recurrent unit demonstrated consistent and smooth motion, effectively maintaining the base roll angle close to the zero reference throughout the simulation.
This is further supported by computing the root mean squared error (RMSE) of the base roll angle~($\theta_z$), over an episode of $H=200$ timesteps~(4s),
\begin{equation}
    \| \theta_z \| = \sqrt{\frac{1}{H}\sum_{t=0}^{H}\hat{\theta}_{z,t}^2} \ . \label{eq: RMSE }
\end{equation}
\Cref{fig:tikz_roll_angle} compares $\| \theta_z \|$ for each robot and \Cref{tab: RMSE(roll) for 32 robots (walk)} reports the average across the test set $\| \theta_z \|_{avg} =  \frac{1}{N_{test}}\sum_{i=1}^{N_{test}}  \| \theta_z \|_i $ for different speed commands, emphasizing the outstanding performance of the GRU policies.

\begin{table} 
\vspace{-0.4cm}
\caption{Walk test: average RMSE of roll angle $\| \theta_z \|_{avg}$~[deg]}
\vspace{-0.2cm}
\begin{center}
\begin{tabular}{|c||c|c|c|}
\hline
$v_{cmd} \ [m/s]$ & GRU & MLP + hist & MLP \\
\hline
    $-0.5$ & \textbf{1.405} & 3.872 & 2.357 \\
    $-0.4$ & \textbf{1.281} & 3.446 & 2.291 \\
    $-0.3$ & \textbf{1.223} & 3.397 & 2.249 \\
    $-0.2$ & \textbf{1.251} & 3.268 & 2.257 \\
    $-0.1$ & \textbf{1.144} & 3.209 & 2.247 \\
    $0$ & \textbf{1.282} & 3.316 & 2.272 \\
    $0.1$ & \textbf{1.122} & 3.268 & 2.349 \\
    $0.2$ & \textbf{1.076} & 3.397 & 2.447 \\
    $0.3$ & \textbf{1.090} & 3.850 & 2.554 \\
    $0.4$ & \textbf{1.402} & 4.149 & 3.194 \\
    $0.5$ & \textbf{1.357} & 4.326 & 2.867 \\
\hline
\end{tabular}
\label{tab: RMSE(roll) for 32 robots (walk)}
\end{center}
\vspace{-0.5cm}
\end{table}

\subsection{Number of Training Robots}
\begin{figure}
    \centering
    \includegraphics[width=\linewidth]{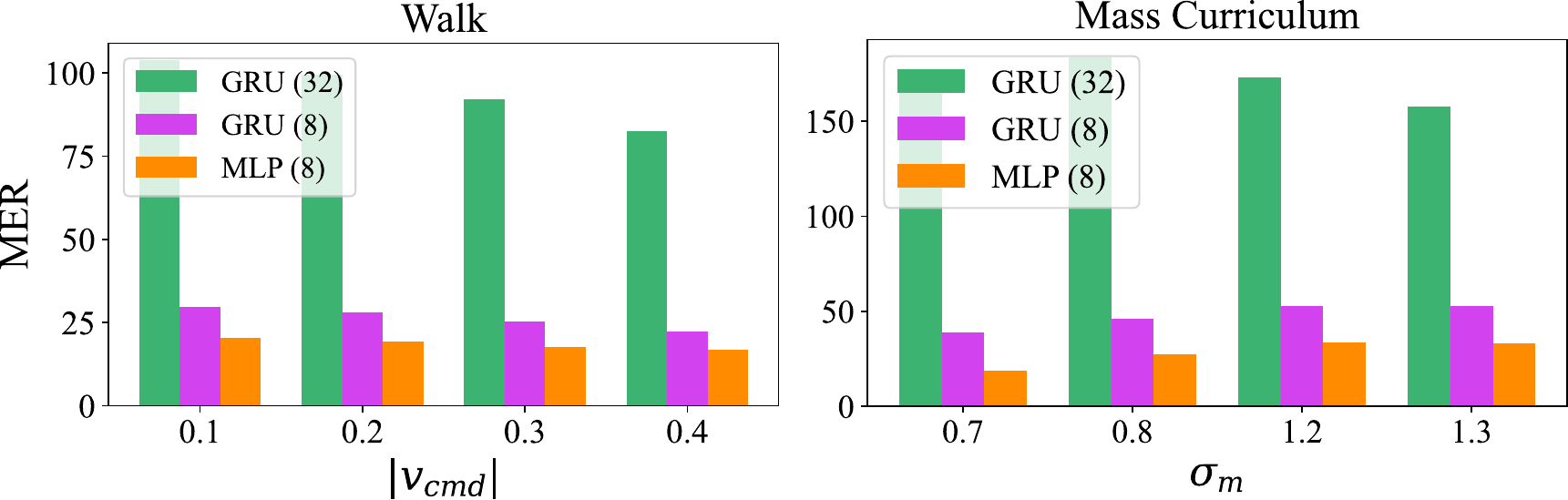}
    \caption{Mean episodic reward (MER) achieved on walk and mass curriculum experiments on different policies: GRU trained on 32 robots, GRU trained on 8 robots and MLP trained on 8 robots.} 
    \label{fig: tikz MER RNN 8 vs 32 train robots}
    \vspace{-0.5cm}
\end{figure}

To evaluate the impact of training set size, we trained 5 GRU policies on a subset of 8 robots from the initial 32.
The performance degradation with fewer training robots, as measured by MER in walk and mass curriculum tests, is summarized in \Cref{fig: tikz MER RNN 8 vs 32 train robots}. 
Policies trained on a smaller set of robots exhibit reduced generalization ability on the test set, and lower robustness to parameter changes, 
primarily due to overfitting. 
This suggests that a sufficiently large and diverse set of robots are required during training to enable the GRU encoder to capture the task distribution and generalize effectively to unseen robots.
While our method necessitates exposure to a sufficient number of robots during training, it is significantly less than what is typically required by current state-of-the-art methods such as~\citet{feng2023genloco}.
An excessively large training set requires extensive computational effort for training.
Furthermore, comparing our policy with MLP policies trained on the same 8 robots highlights that the GRU module is essential for achieving higher generalization from a smaller training set, as depicted in \Cref{fig: tikz MER RNN 8 vs 32 train robots}.

\subsection{Meta-episode Length K}
\label{result:meta-episode}
\begin{figure} [t]
    \centering    
    \includegraphics[width=0.9\linewidth]{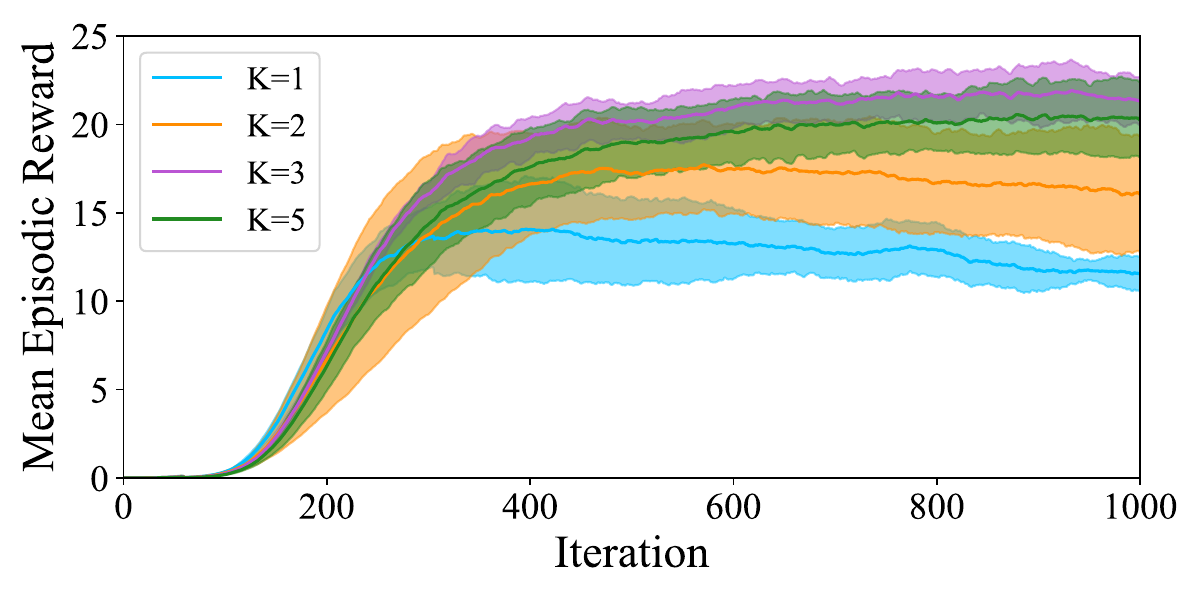}
    \vspace{-0.15cm}
    \caption{Mean and standard deviation of reward curves during training for different meta-episode lengths K, over 5 different seeds.} 
    \label{fig: tikz ablation study Meta-episode length}
\end{figure}

In this section, we provide an ablation study on the meta-episode length $K$.
For each value of $K$, 5 policies are trained with different random seeds.
The learning curves in \Cref{fig: tikz ablation study Meta-episode length} reveal that lower values of $K$ either result in a decreasing reward trend or a high variance between training runs.
Conversely, preserving a multi-episodic memory (i.e. $K > 2$) leads to stable training curves and higher final rewards.
This experiment highlights the importance of maintaining the GRU hidden state across episodes, as also noted in prior research~\cite{duan2016rl}, where the meta-episode length $K$ serves as a problem-dependent hyperparameter.

\section{Hardware results}
\label{sec:hw}
To assess the effectiveness of the learned controller,
we deployed it on three distinct commercial quadruped robots: \textit{Go1},  \textit{Go2}, and  \textit{Aliengo} from  \textit{Unitree}~\cite{unitree}.
Notably, the training robots were generated by randomly scaling the parameters of \textit{Go1} and \textit{Aliengo}, but none of the actual robots were specifically encountered during the training phase. 
Dynamic randomization was employed during training to facilitate robust real-world performance. 
This included randomizing the terrain friction coefficient, gravity vector $g$, and training on uneven terrain~\cite{lee2020learning}. 
Additionally, we randomized actuator delay, added noise to several observations and incorporated impulse perturbations during training.
We note that dynamics randomization is crucial for successful sim-to-real transfer with recurrent policies, to counteract the tendency of RNNs to overfit to simulation dynamics~\cite{siekmann2020learning}. 
As demonstrated in the supplementary video and \Cref{fig: hardware deployment}, the universal policy trained with our method successfully controlled locomotion across the three commercial quadrupeds. 
This achievement is attributed to the RNN's capability to conduct implicit system identification through its hidden state.

\begin{figure}
    \centering
    \includegraphics[width=\linewidth]{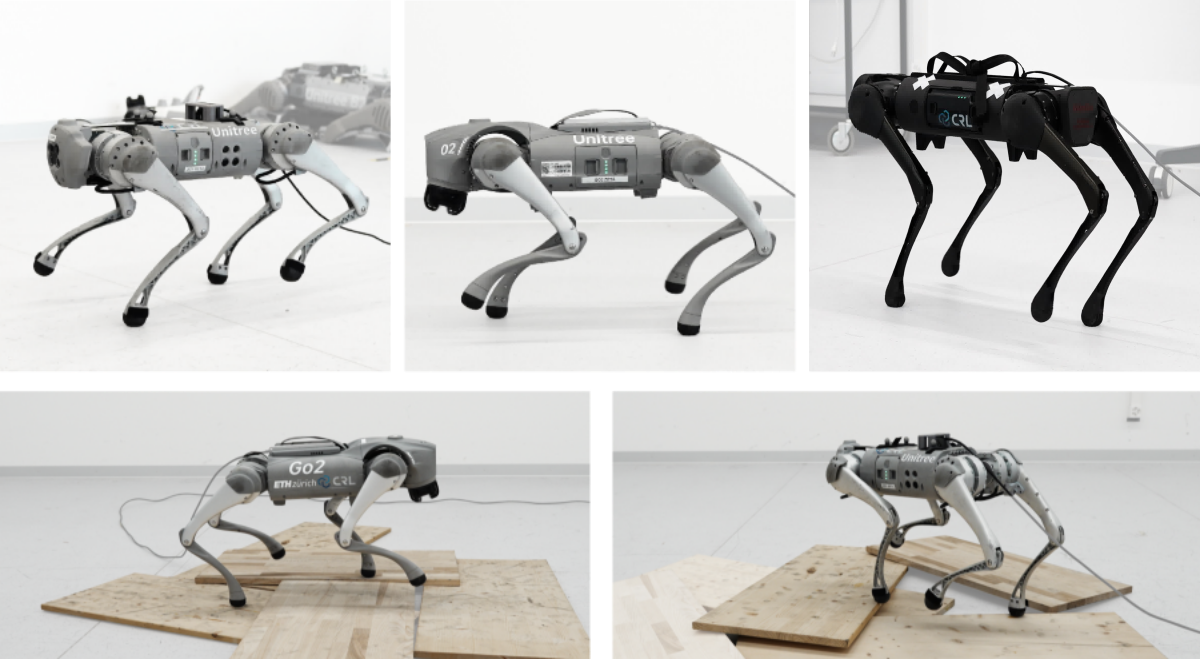}
    \caption{Snapshots of the  \textit{Unitree Aliengo} \textbf{(top right)},  \textit{Unitree Go2} \textbf{(top center, bottom left)} and  \textit{Unitree Go1} \textbf{(top left, bottom right)} successfully performing trot motions, guided by our single universal policy.} 
    \label{fig: hardware deployment}
    \vspace{-0.5cm}
\end{figure}

\section{Conclusions and Future Work}
In this work, we introduce a meta-RL approach to train a universal locomotion controller for quadrupeds, which enables zero-shot transfer to diverse legged robotic platforms with varying physical and kinematic properties.
By employing working-memory meta-RL, our method offers a straight-forward solution to the challenge of robot-agnostic locomotion control, while achieving higher sample efficiency. 
In the simulation experiments, our method showed generalization from a small number (32) of training robot morphologies.
Furthermore, our hardware experiments demonstrate that our universal policy achieves zero-shot transfer in real-world to three different quadruped platforms,
despite them not being seen during training.
The main limitation of our approach is its applicability primarily to robots sharing a morphological template with 12 degrees of freedom.  
In the future, more flexible policy architectures, such as GNN or transformers, can be explored to remove the constraint on the robot morphological template and control quadrupeds with different leg designs, such as the x-shaped leg or three-link leg designs.

\bibliographystyle{IEEEtranN}
\bibliography{root.bib}

\end{document}